\documentclass[sigconf,9pt]{acmart}
\AtBeginDocument{%
  }

\usepackage[utf8]{inputenc}
\usepackage[bottom]{footmisc}
\usepackage{subcaption}
\usepackage{graphicx} 
\usepackage[most]{tcolorbox}
\usepackage[table,xcdraw]{xcolor} 
\usepackage{algorithmic}
\usepackage{graphicx}
\usepackage{textcomp}
\usepackage{multirow}
\usepackage{xcolor}
\usepackage{enumitem}
\usepackage{xspace}
\usepackage{pifont}
\usepackage{booktabs}
\usepackage{url}
\usepackage{caption}    

\newcommand{\Method}{HyperLiDAR\xspace} 
\newcommand{\xiaofan}[1]{{\color{black}#1}}
\newcommand{\YeTian}[1]{{\color{black}#1}}

\AtBeginDocument{%
  }

\copyrightyear{2026}
\acmYear{2026}
\setcopyright{cc}
\setcctype{by}
\acmConference[DAC '26]{63rd ACM/IEEE Design Automation Conference}{July 26--29, 2026}{Long Beach, CA, USA}
\acmBooktitle{63rd ACM/IEEE Design Automation Conference (DAC '26), July 26--29, 2026, Long Beach, CA, USA}
\acmDOI{10.1145/3770743.3804248}
\acmISBN{979-8-4007-2254-7/2026/07}

\begin{document}

\title{HyperLiDAR: Adaptive Post-Deployment LiDAR Segmentation via Hyperdimensional Computing\\
}


\author{
  Ivannia Gomez Moreno\textsuperscript{*}$^{1}$, Yi Yao\textsuperscript{*}$^{1}$, Ye Tian\textsuperscript{*}$^{1}$, Xiaofan Yu$^{2}$, Flavio Ponzina$^{3}$, Michael Sullivan$^{1}$, Jingyi Zhang$^{1}$, Mingyu Yang$^{4}$, Hun Seok Kim$^{4}$ and Tajana Rosing$^{1}$\\
  $^{1}$ UCSD, $^{2}$ UC Merced, $^{3}$ SDSU, $^{4}$ UMich}
\email{{ivgomezmoreno, yiy054, yet002, mis016, jiz290, tajana}@ucsd.edu}
\email{xiaofanyu@ucmerced.edu, fponzina@sdsu.edu, {mingyuy,hunseok}@umich.edu}

\thanks{\textsuperscript{*}These authors contributed equally to this research. Ye Tian is corresponding author.}



\begin{abstract}
LiDAR semantic segmentation plays a pivotal role in 3D scene understanding for edge applications such as autonomous driving. 
\xiaofan{However, significant challenges remain for real-world deployments, particularly for on-device post-deployment adaptation. Real-world environments can shift as the system navigates through different locations, leading to substantial performance degradation without effective and timely model adaptation. Furthermore, edge systems operate under strict computational and energy constraints, making it infeasible to adapt conventional segmentation models (based on large neural networks) directly on-device.} To address the above challenges, we introduce \Method, the first lightweight, post-deployment LiDAR segmentation framework based on Hyperdimensional Computing (HDC).
\xiaofan{The design of \Method fully  leverages the fast learning and high efficiency of HDC, inspired by how the human brain processes information. To further improve the adaptation efficiency, we identify the high data volume per scan as a key bottleneck and introduce a buffer selection strategy that focuses learning on the most informative points.}
We conduct extensive evaluations on two state-of-the-art LiDAR segmentation benchmarks and two representative devices.
Our results show that \Method outperforms or achieves comparable adaptation performance to state-of-the-art segmentation methods, while achieving up to a 13.8$\times$ speedup in retraining.
\end{abstract}


\maketitle

\section{Introduction}
\label{sec:intro}

%

\xiaofan{LiDAR sensors emit laser pulses to measure the distance to surrounding objects, producing a three-dimensional representation of the environment known as a point cloud. This capability for 3D perception is widely recognized as a key enabler of next-generation spatial AI~\cite{spatialai}. LiDAR is broadly used in real-world edge applications, including autonomous driving~\cite{piroli2024label, li2024efficient}, structural analysis~\cite{ni2024quantitative, de2024complete}, and environmental studies~\cite{li2024extracting, tan20243d, bilodeau2024enhancing}. In this paper, we mainly focus on autonomous driving, although the proposed techniques can be readily extended to other scenarios.}

\xiaofan{LiDAR semantic segmentation evaluates a spatial AI system’s ability to understand 3D environments from LiDAR data. The goal is to assign an accurate category label (e.g., car or pedestrian) to each individual point in the point cloud. While existing neural network models have achieved promising segmentation accuracy in offline training~\cite{zhu2021cylindrical}, a significant gap remains in deploying these techniques in the real world. In particular, real-world environments and their LiDAR scan patterns can change as the vehicle moves through different locations, e.g., from urban to rural areas or other unseen scenarios, leading to accuracy drops in fixed segmentation models~\cite{michele2024saluda}.}
To handle such changes, fast and timely post-deployment adaptation is needed to continuously retrain the model on new LiDAR scans. 
Achieving this goal presents two key challenges:
\textit{First,} conventional models such as neural networks suffer from catastrophic forgetting, a phenomenon in which learning new data disrupts previously acquired knowledge~\cite{mccloskey1989catastrophic}. This makes these models unsuitable for continuous adaptation in real-world streaming environments.
\textit{Second,} an autonomous driving system typically operates on resource-constrained hardware, whose storage and computing capabilities are significantly lower than those of desktop-level GPUs. More efficient methods are required to adapt directly on the edge device.

Although both challenges are critical, existing work on LiDAR point cloud segmentation has mainly focused on solving the first one.
One line of research has explored cross-domain adaptation in autonomous vehicles~\cite{wang2023cross, chen2018domain, akhauri2021improving, li2022cross, michele2024saluda}, aiming to leverage prior knowledge to manage domain shifts in real-world applications. 
Other recent works have attempted to improve the efficiency of LiDAR segmentation models, focusing on training~\cite{xu2020squeezesegv3, cheng2023transrvnet, cheng2022cenet, li2024tfnet, chen2021rangeseg} or inference~\cite{li2023less, wang2024segnet4d}.
However, all of these models are based on deep neural networks, which are too resource-intensive to train on edge devices.
In short, no existing work addresses both challenges required for effective and efficient adaptation after deployment.

In this paper, we propose \Method, a novel Hyperdimensional Computing (HDC)-based LiDAR segmentation approach that simultaneously \xiaofan{offers effective and lightweight post-deployment adaptation on edge systems}. 
%
Inspired by how the human brain processes information, Hyperdimensional Computing~\cite{kanerva2009hyperdimensional} enables fast, lightweight learning by encoding raw sensory inputs into high-dimensional (e.g., 10K) low-precision vectors, then using only element-wise additions and multiplications in training and inference. This makes HDC far more efficient for resource-constrained edge devices than training deep neural networks~\cite{kim2018efficient, ponzina2024glucosehd}, and an ideal solution for post-deployment adaptation.
\xiaofan{Specifically, \Method first encodes each point in a LiDAR scan into a high-dimensional, low-precision vector, called a \textit{hypervector}. Fast and efficient adaptation is then performed by updating a representative hypervector for each segmentation class (e.g., cars, pedestrians), without requiring any gradient descent.}
To improve adaptation efficiency on large LiDAR scans with 50K–60K points, \Method further introduces a buffer selection strategy to quickly and adaptively select the worst-performing samples from previous scans. Using this strategy, \Method can retrain hypervectors with only 5\% of data while maintaining accuracy comparable to that of using all point clouds, \xiaofan{significantly increasing the number of scans that can be processed per second.} 
In summary, the main contributions of this work are as follows:
\begin{itemize}
\item We propose \Method, a novel post-deployment adaptive learning framework based on HDC for LiDAR semantic segmentation. \xiaofan{To the best of our knowledge, \Method is the first HDC-based framework designed for post-deployment LiDAR segmentation.}

\item We design a novel buffer selection strategy that incorporates historically underperforming samples during the adaptive update process. \xiaofan{The buffer selection strategy is a key component of \Method, addressing the main bottleneck of large LiDAR data volume and achieving similar adaptation performance using just 5\% of the data.}  

\item 
Extensive evaluations on two widely used datasets (Semantic-KITTI \cite{behley2019semantickitti} and nuScenes \cite{caesar2020nuscenes}) and two representative hardware platforms demonstrate that, compared with state-of-the-art baselines, \Method achieves superior performance while delivering up to $13.8{\times}$ acceleration.
\end{itemize}
\section{Related Works}
\label{sec:related-works}

\subsection{Domain Adaptation of LiDAR Segmentation} 

Domain adaptation tackles the challenges posed by dynamic data by developing models that adapt to the vast and continually evolving environments \cite{spatialai}. This encompasses domain shifts in target distribution and covariance \cite{michele2024saluda}.
In the traditional pipeline, models develop general features by incorporating additional modules that align features from various domains. Akhauri et al. \cite{akhauri2021improving} emphasize the importance of generalizing features across domains to enhance transfer learning in the context of steering angle prediction. Chen et al. \cite{chen2018domain} calculated the difference between domains to facilitate the alignment of features from differing distributions. DAYOLO \cite{li2022cross} introduces a regularization and adaptation module designed to produce domain-invariant representations. Likewise, Wang et al. \cite{wang2023cross} embed a domain adaptation module within traditional segmentation architectures to minimize discrepancies in feature distributions between the source and target domains. SALUDA \cite{michele2024saluda} introduces an unsupervised auxiliary task that learns an implicit underlying surface representation on both source and target data.
However, these methods generally require substantial computational and memory resources, and thus still face significant bottlenecks when adapted for deployment on resource-constrained edge devices.

\subsection{Efficient LiDAR Segmentation}

Processing large-scale point clouds has been widely recognized as a key bottleneck in LiDAR segmentation~\cite{li2023less}.
To improve inference efficiency, methods such as SqueezeSegV3~\cite{xu2020squeezesegv3}, TransRVNet~\cite{cheng2023transrvnet}, CENET~\cite{cheng2022cenet}, TFNet~\cite{li2024tfnet}, and RangeSeg~\cite{chen2021rangeseg} convert LiDAR scans into image-like representations. This reduces computational load, as the number of pixels is much smaller than the number of 3D points in a scan. However, training convolutional models on these representations still requires considerable computational resources, making them less practical for resource-constrained edge devices.
Other works have aimed to improve the training efficiency of LiDAR segmentation models, often using sparse networks.
Less is More~\cite{li2023less} and SegNet4D~\cite{wang2024segnet4d} adopt sparse convolutions or sparse encoder-decoder architectures to significantly reduce parameter counts and model size compared to prior approaches.
However, they lack support for post-deployment adaptation in dynamic environments due to their high computational requirements.
In contrast to prior work, \Method delivers both accurate and efficient adaptation in post-deployment environments, addressing a key gap for real-world deployment.


\begin{figure*} 
    \centering
    \includegraphics[width=0.9\textwidth]{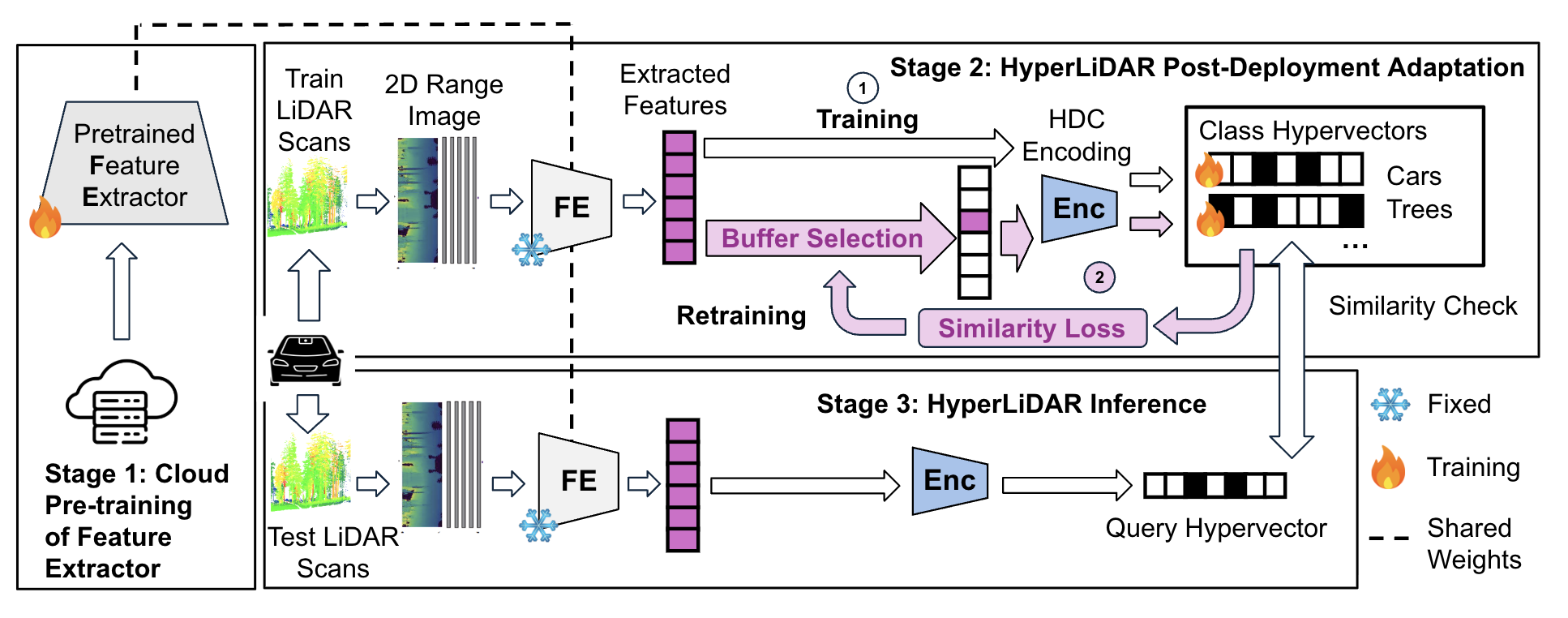} 
    \vspace{-2mm}
    \caption{\small{The pipeline of \Method including three stages. The pre-deployment training is performed in the cloud. The post-deployment adaptation and inference are performed on edge devices, taking streaming LiDAR scans as inputs.}}
    \label{fig:HyperLiDAR}
    \vspace{-2mm}
\end{figure*}

\section{Background}
\subsection{Problem Definition}

We consider the problem of adaptive post-deployment LiDAR semantic segmentation. This task is critical for scene understanding in real-world applications like autonomous driving, requiring the assignment of a semantic label (e.g., cars, pedestrians) to every point in a 3D point cloud scan.
Once a model is initially trained and deployed, the environment changes constantly. For instance, the input point cloud distribution can continuously vary due to environmental (e.g., night/day, rain/fog/sun) and urban (e.g., city/rural) factors. This can lead to a drop in the model’s accuracy ~\cite{michele2024saluda}. To manage these dynamic changes, the model requires post-deployment adaptation (or retraining) on new, streaming LiDAR scans. 

\xiaofan{Following prior post-deployment training works~\cite{Llisterri2022OnDevice, lin2024ondevicetraining256kbmemory,10.1145/3526241.3530373}, we formulate the concrete problem as a three-stage process.  
\textit{1.)} Pre-deployment training: We first pretrain the model offline using a pre-collected LiDAR dataset $\{ \mathcal{X}_{pre}, y_{pre} \}$, where $\mathcal{X}_{pre}$ denotes raw point cloud inputs and $y_{pre}$ denotes the corresponding segmentation labels.  
\textit{2.)} Post-deployment adaptation: After deployment, the pretrained model is adapted on-device using streaming real-world data $\{ \mathcal{X}_{post}, y_{post} \}$. Note that the pre-deployment and post-deployment datasets are completely disjoint, i.e., all samples encountered during adaptation are unseen during offline training. In addition, the data distributions differ, $p(\mathcal{X}_{pre}, y_{pre}) \neq p(\mathcal{X}_{post}, y_{post})$, reflecting real-world domain shifts.  
\textit{3.)} Evaluation: Finally, we evaluate the adapted model on a universal test set $\{ \mathcal{X}_{test}, y_{test} \}$.}


\subsection{HDC Background}\label{sec:HDC Background}

In the problem defined above, Hyperdimensional Computing (HDC) offers a promising solution, as it excels at lightweight and efficient adaptation.
HDC is inspired by the neuroscience community, seeking to emulate human brain functioning~\cite{kanerva2009hyperdimensional}. HDC operates by mapping samples into high-dimensional vectors, known as \textit{hypervectors}, which often consist of thousands of dimensions. The fundamental idea behind HDC is that intricate patterns within the original space can potentially be linearly separable when projected into a sparse, high-dimensional space~\cite{moin2021wearable}.
Learning in high-dimensional space can be achieved through simple and highly parallelizable operations. As a result, HDC bypasses the costly gradient descent operations used in deep neural networks, enabling lightweight post-deployment learning.

\textbf{Encoding}: The first and key step in HDC is encoding, which maps features extracted from input samples into high-dimensional, low-precision vectors, known as \textit{hypervectors}.
Previous studies have explored various encoding techniques, including random projection~\cite{thomas2021theoretical}, ID-level encoding~\cite{karunaratne2020memory}, and spatiotemporal encoding~\cite{moin2021wearable}.




\textbf{Training}: During the initial training epoch,
HDC generates a representative class hypervector for each class by adding the hypervectors of all samples within that class~\cite{thomas2021theoretical}.
In this way, a hypervector closely related to all the samples in that class is formed and stored. 
Suppose $\{(x_{i},y_{i})\}$ is a set of labeled samples from class $j$, the class hypervector representing class $c_j$ is computed as
$c_{j} = \bigoplus_{i : y_{i} = c_{j}} \phi(x_{i})$. $\bigoplus$ denotes a simple element-wise addition.
Notice that training an HDC classifier involves only hypervector additions, resulting in significantly lower computational complexity compared to neural network training.

\textbf{Inference}: During inference, the cosine similarity between the query input $x_q$ and each class hypervector is computed. The predicted class $\hat{y}$ is assigned to the one whose hypervector has the highest similarity score with the query.
Similar to HDC training, the inference stage involves lightweight vector similarity computations, enabling energy-efficient on-device inference.

\section{\Method}
\label{sec:method}
In this section, we introduce \Method, a hyperdimensional computing–based framework for LiDAR point cloud segmentation that transforms fixed offline models into efficient, self-adapting systems, enabling low-latency training suitable for real-world deployment.


\subsection{Overview}







Fig.~\ref{fig:HyperLiDAR} outlines the full deployment pipeline of \Method. As described in the problem definition, it consists of three stages: \textit{1.)} pre-deployment training in the cloud, \textit{2.)} post-deployment adaptation on edge, and \textit{3.)} post-deployment inference on edge. 
In the first stage, we pretrain the model using large-scale LiDAR segmentation data in the cloud to obtain a feature extractor that is both efficient and universal.
This extractor is then frozen and used in conjunction with the HDC model in the subsequent stages. A generalized feature extractor allows \Method to extract key features from a large variety of environments, ultimately supporting the training of an efficient HDC classifier.
To ensure computational efficiency, we adopt a projection-based feature extractor~\cite{cheng2022cenet}, which converts 3D point clouds into 2D image representations before encoding. This approach has been shown to be significantly more efficient for downstream processing~\cite{cheng2022cenet,xu2020squeezesegv3,chen2021rangeseg}.

The second and third stages are the key focus of \Method, i.e., post-deployment adaptation and inference. To achieve these goals, \Method incorporates two key designs: First, \Method leverages the inherently lightweight nature of HDC in adaptation and inference pipelines. Second, recognizing that the primary efficiency bottleneck lies in the large number of data points in each LiDAR scan, \Method introduces a buffer selection strategy that prioritizes training on the ``hardest'' points.
This mechanism allows the model to achieve comparable results using only 5\% of the full dataset in each training stage, significantly improving efficiency on edge devices. 
In the following sections, we will provide more details on each design.

\subsection{HDC-based Training and Inference}

\Method enables post-deployment adaptation to new scenarios through an HDC-based training and inference pipeline, as illustrated in Stages 2 and 3 of Fig.~\ref{fig:HyperLiDAR}.
This is achieved by combining a lightweight and generalized feature extractor, pre-trained offline in the cloud, with a lightweight and trainable HDC module. Once pretrained (as described in Stage 1), the feature extractor remains frozen during deployment. This design allows the system to leverage high-quality, general-purpose feature representations without incurring any backpropagation overhead on the edge device.


\textbf{\Method Encoding:} In both training and inference stages, encoding is the first step to map the features extracted by the frozen encoder into hypervectors.
In \Method, we employ the HDnn encoding~\cite{dutta2022hdnn, yu2024lifelong}, which combines a lightweight pretrained feature extractor with random projection encoding.
HDnn is essential for handling complex data structures like images and LiDAR scans, enabling effective feature extraction and improved accuracy.
Formally, let $x$ represent the raw point clouds from LiDAR scans, and let $z = f(x)$ denote the features extracted by a pretrained feature extractor, where $z$ has dimension $p$. 
Random projection~\cite{thomas2021theoretical} is then applied to these features by multiplying $z$ element-wise with a random projection matrix $M \in \mathbb{R}^{p\times d}$, whose entries are drawn from a normal distribution $\mathcal{N} (0,1)$. The resulting vector $hv \in \mathbb{R}^d$ is normalized using $L_2$ normalization.
Applying this process to all training samples produces a set of high-dimensional hypervectors $hv_2 \in \mathbb{R}^d$. 
Finally, all dimensions of the hypervector are binarized into a bipolar format ($\{-1,1\}$) using the sign function, resulting in a low-precision representation.
Mathematically, the encoding process can be expressed as:
\begin{subequations}
\begin{align}
    z &= f(x) \\
    hv &= z \times M \\
    \quad hv_2 &= \frac{hv}{\mathrm{max}(\left \| hv \right \|_2, \epsilon)}\\
    \phi(x) &= \mathrm{sign}(hv_2)
\end{align}
\end{subequations} 
where $\phi(x)$ denotes the encoded hypervector.

\textbf{\Method Inference:} The inference process is shown in Stage 3 of Fig. \ref{fig:HyperLiDAR}. After feature extraction, each point in a new LiDAR scan is classified using a standard HDC pipeline, adapted for large-scale LiDAR data.
For a query point of $x_q$, segmentation labels are assigned by comparing the encoded hypervector of $x_q$ with all class hypervectors as follows:
\begin{equation}
\hat{y} = \text{argmax}_{j} \cos \left (\phi(x_q), c_{j} \right ).
\end{equation} 


\textbf{HyperLiDAR Post-Deployment Adaptation:} During post-deployment adaptation, only the class hypervectors are updated. Each class hypervector $c_j$ represents a semantic class for the LiDAR scans, e.g., cars, trees. \Method includes $e$ additional retraining epochs to refine these class hypervectors beyond the initial pass. However, the total number of epochs is kept limited to maintain training efficiency on edge devices.
During each retraining phase of \Method, class hypervectors are updated to maximize intra-class data similarity and inter-class distance. To achieve this, the following class updates are performed for each misclassified input sample $(x_i, y_i)$:
\begin{subequations}
\begin{align}
c_j &\leftarrow c_j + \phi(x_i) \\
c_k &\leftarrow c_k - \phi(x_i)
\end{align}
\end{subequations}
where $c_j$ is the ground-truth class hypervector for label $y_i$ and $c_k$ is the hypervector of the mispredicted class, with $j\neq k$.

\subsection{Buffer Selection Strategy} 


LiDAR scans contain a large volume of data, with each scan having 50K to 60K points in SemanticKITTI~\cite{behley2019semantickitti}. While HDC is a lightweight and efficient learning approach that avoids costly gradient descent, the process of encoding this massive amount of data into high-dimensional vectors still creates a major computational bottleneck, consuming nearly 50\% of the retraining time per batch.

To address this, we leverage a key property of HDC: its ability to learn effectively and converge quickly even with limited data~\cite{yang2024fsl}. By designing a novel buffer selection strategy, \Method can significantly reduce the number of data points processed during retraining, thereby alleviating the encoding bottleneck and dramatically reducing adaptation latency without sacrificing performance.

This strategy is an efficiency-driven coreset selection optimally designed for HDC's lightweight training mechanism. Unlike computationally heavy coreset methods like Gradient Coresets or k-Center Greedy, which demand complex gradient or distance calculations, our method is simple, fast, and highly effective. Our design is centered on selecting the most informative subset of data by prioritizing the "hardest" samples in the current iteration. To achieve this, \Method first performs a full data update using all data in the initial post-deployment epoch. In subsequent retraining iterations, a new buffer is selected at the end of each epoch based on the latest performance updates, enabling the model to adapt to difficult cases without costly full data passes. The detailed process is as follows:

\textbf{Initial Post-Deployment Epoch}: 
After the pretrained feature extractor is deployed on device, the first pass of post-deployment training is performed on the full dataset. During this initial pass, each point is classified, and the classification loss is computed for each point using a custom cosine-similarity-based loss function.
To quantify the misclassification cost, we define a loss function inspired by multi-class perceptron learning~\cite{beygelzimer2019bandit} as follows:
\begin{equation}
L(\phi; x_i, y_i) = \begin{cases}
0 & \text{if } y_i = \hat{y}_i \\
\cos \left (\phi(x_i), c_k \right) - \cos \left (\phi(x_i), c_j \right) & \text{if } y_i \neq \hat{y}_i
\end{cases}
\end{equation}
Here, $\phi(x_i)$ denotes the encoded hypervector of input $x_i$, $c_j$ denotes the ground-truth class hypervector associated with label $y_i$, and $c_k$ denotes the hypervector of the mispredicted class $\hat{y}_i$. Intuitively, the loss captures how far a misclassified prediction deviates from the ground-truth semantic class.

\textbf{Buffer Selection Retraining Epochs:} 
In subsequent retraining passes, we apply the buffer selection strategy to significantly reduce the data volume. By composing the buffer with both ``hard'' and ``random'' samples, the model can efficiently focus its learning efforts where they are needed most while still maintaining its ability to generalize. This design ensures that the model learns from the most valuable data points without the computational overhead of processing every point in the LiDAR scan. 
We select a small subset comprising k\% of the features for training, composed of two equally important parts:

\textit{Hard Samples:} 
Half of the considered features are selected from data with the highest classification loss from the previous retraining pass.
These features represent the ``hard'' samples that the model struggles to classify correctly. Prioritizing these samples during retraining helps maximize learning efficiency by directly targeting areas of uncertainty and misclassification. We select the top $0.5k\%$ of samples with the highest losses to prioritize harder samples.

\textit{Random Samples:} The second half of the retraining data subset is selected randomly from the remaining data within the current batch. This ensures diversity in the retraining data and helps prevent overfitting to only the hard samples. 
Introducing random samples is important for maintaining the generalization of the model under various conditions. This can prevent it from focusing only on difficult samples, thereby hindering its performance in other typical scenarios.


\textbf{Loss Update:} After each retraining pass on the selected buffer, the classification loss for the samples in the buffer is updated. The loss for any data not included in the current buffer remains unchanged, preserving its value for potential selection in the next pass. In each subsequent epoch, the model re-evaluates which samples are currently the most challenging based on these stored loss values, enabling it to continuously adapt to new difficult cases without incurring the computational overhead of a full data pass.

\section{Evaluation}
\label{sec:evaluation}

\subsection{Experimental Setup} 





\textbf{\textit{Dataset.}}
We evaluate HyperLiDAR and baseline methods on two large-scale autonomous driving benchmarks, SemanticKITTI~\cite{behley2019semantickitti} and nuScenes~\cite{caesar2020nuscenes}, to validate its superiority under realistic online deployment scenarios. SemanticKITTI \cite{behley2019semantickitti} is a LiDAR-based semantic segmentation benchmark comprising 43,551 scans from 22 driving sequences. nuScenes \cite{caesar2020nuscenes} provides 1,000 urban driving scenes, covering various weather and traffic conditions. 
For SemanticKITTI, we select 8 scenes for pre-training, 2 scenes for post-deployment, and use one new scene for final evaluation following prior work~\cite{cheng2022cenet,puy2024three,puy2023using,zhu2021cylindrical}. For nuScenes, we split the training set with an 8:2 ratio, resulting in 560 scenes for pre-training and 140 for post-deployment, while the 150 unseen scenes are reserved for evaluation. The scenes set in each stage are disjoint.



\textbf{\textit{Baselines. }}
We compare with the following SOTA baselines and \Method without buffer selection. 
 \textit{a.)} \textbf{CENET}~\cite{cheng2022cenet} projects 3D spherical coordinates into 2D multimodal representations and applying CNN-based feature extraction. Through the use of tailored activation functions, CENET achieves competitive accuracy and real-time inference on large-scale benchmarks, and serving as a strong baseline for LiDAR segmentation.
 \textit{b.)} \textbf{SALUDA}~\cite{michele2024saluda} introduces an unsupervised auxiliary task that learns a shared implicit surface representation across source and target domains, aligning features from different sensors and driving conditions and achieving advanced performance in scene-adaptive LiDAR semantic segmentation.
\textit{c.)} \textbf{HyperLiDAR$-\textrm{full}$}: We also compare with HyperLiDAR$-\textrm{full}$, which executes the HyperLiDAR pipeline without buffer selection, to validate the effectiveness of the proposed buffer selection strategy and its significant efficiency improvements.


\textit{\textbf{Implementation Details.}}
The hyperparameters are chosen empirically to balance accuracy, runtime, and the computational constraints of edge devices. Specifically, considering the limitations of input size and resources, we use a batch size of $B=6$, an HDC dimensionality of $D=10{,}000$, and a feature-extractor output dimension of $p=128$. For buffer selection, we set the ratio to $k=5$, i.e., only $5\%$ of the features (comprising $2.5\%$ hard samples and $2.5\%$ randomly sampled ones) are used for retraining.

\textbf{\textit{Metrics.}}
We evaluate the adaptability of each method using mean Intersection over Union (mIoU), a standard accuracy metric for LiDAR segmentation~\cite{cheng2022cenet, zhu2021cylindrical}.
mIoU measures the average overlap between predicted and ground truth segments across all classes.
To evaluate training efficiency, we use the average FPS per LiDAR scan and conduct a comprehensive assessment on two representative platforms with different resource levels:
\begin{itemize}
    \item \textbf{NVIDIA RTX 4090 GPU}~\cite{4090}, a powerful platform for fair comparison across methods, including baselines not optimized for edge devices.
    \item \textbf{FSL-HDnn}~\cite{yang2024fsl}, a custom HDnn ASIC. Although FSL-HDnn is not designed for LiDAR segmentation, we adapt it to accelerate \Method, providing a conservative lower bound acceleration for specialized edge hardware. The FPS is estimated based on the cycle number of each operation.

\end{itemize}

\begin{table}[h]
\centering
\resizebox{0.47\textwidth}{!}{
\begin{tabular}{llc*{7}{c}}
\toprule
\multirow{2}{*}{\textbf{Dataset}} & \multirow{2}{*}{\textbf{Model}} & \multirow{2}{*}{\textbf{mIoU (\%)}}
& \multicolumn{6}{c}{\textbf{IoU (\%)}} \\
\cmidrule(lr){4-9}
 &  &  & car & bicycle & motorcycle & truck & pedestrian & terrain \\
\midrule
\multirow{4}{*}{SemanticKITTI \cite{behley2019semantickitti}}
& CENET \cite{cheng2022cenet}              & \textbf{56.9} & 94.9 & 32.3 & 33.8 & \textbf{75.4} & 53.3 & \underline{75.9} \\
& SALUDA \cite{michele2024saluda}          & 49.1 & 93.3 & 1.0 & 30.5 & 16.9 & 40.0 & \textbf{90.3} \\
& HyperLiDAR (Ours)                        & 54.0 & \textbf{95.2} & \underline{21.4} & \underline{50.7} & 61.9 & \underline{59.6} & 66.7\\
& HyperLiDAR-full (Ours)                   & \underline{55.7} & \textbf{95.2} & \textbf{39.8} & \textbf{52.9} & \underline{62.2} & \textbf{60.9} & 68.9 \\
\midrule
\multirow{4}{*}{nuScenes \cite{caesar2020nuscenes}}
& CENET \cite{cheng2022cenet}              & 54.5 & 86.8 & \underline{12.8} & 57.9 & 60.7 & 49.0 & 61.0 \\
& SALUDA \cite{michele2024saluda}          & \underline{65.2} & 80.5 & 5.29 & 58.0 & 60.8 & \textbf{60.8} & 70.8 \\
& HyperLiDAR (Ours)                        & 61.6 & \underline{91.8} & 7.3 & \underline{65.3} & \textbf{78.8} & 37.7 & \textbf{72.3} \\
& HyperLiDAR-full (Ours)                   & \textbf{67.3} & \textbf{92.0} & \textbf{34.7} & \textbf{67.7} & \textbf{78.8} & \underline{50.9} & \underline{72.0} \\
\bottomrule
\end{tabular}}
\centering
\caption{Performance on SemanticKITTI and NuScenes.
}
\vspace{-0.3in}
\label{tbl:accuracy}
\end{table}

\begin{figure*}[t]
\centering
\begin{minipage}[t]{0.46\textwidth}
  \centering
  \begin{minipage}[t]{0.49\textwidth}
    \centering
    \includegraphics[width=\linewidth]{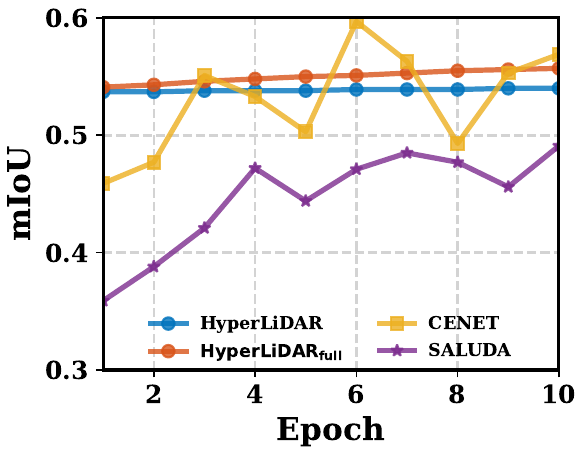}
  \end{minipage}
  \begin{minipage}[t]{0.49\textwidth}
    \centering
    \includegraphics[width=\linewidth]{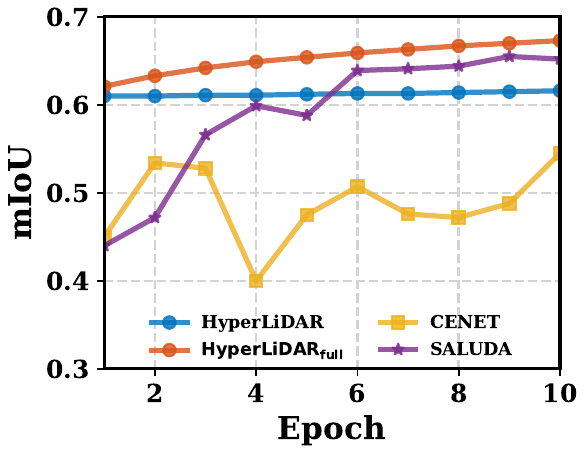}
  \end{minipage}
  \captionof{figure}{Training convergence in terms of mIoU versus epoch on SemanticKITTI (left) and nuScenes (right).}
  \label{fig:epoch_iou}
\end{minipage}
\hfill
\begin{minipage}[t]{0.24\textwidth}
  \centering
  \includegraphics[width=\linewidth]{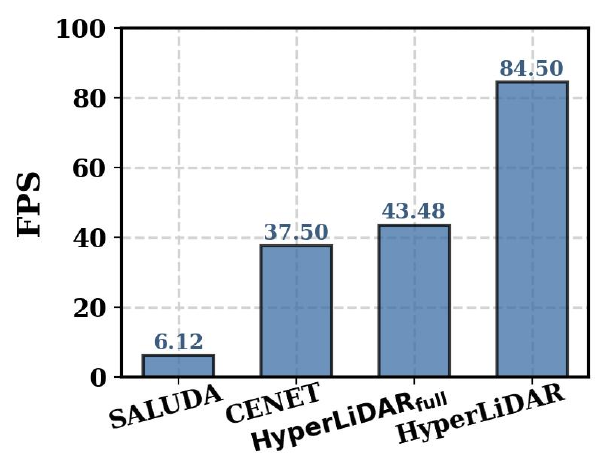}
  \captionof{figure}{Average FPS on NVIDIA RTX 4090.}
  \label{fig:ALL_FPS}
\end{minipage}
\hfill
\begin{minipage}[t]{0.25\textwidth}
  \centering
  \includegraphics[width=\linewidth]{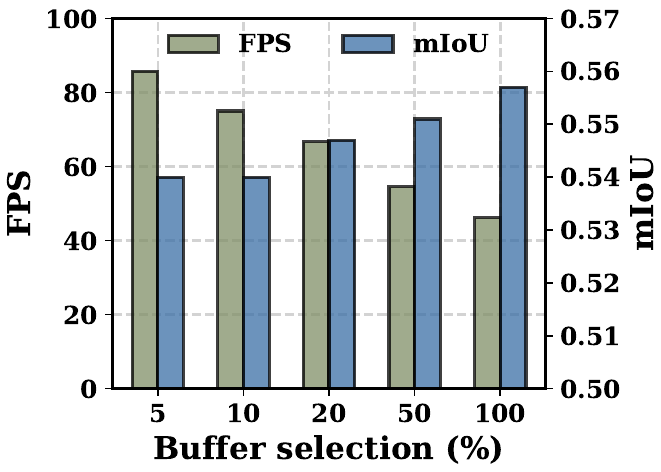}
  \captionof{figure}{Impact of buffer selection size for \Method.}
  \label{fig:buffer_selection}
\end{minipage}
\end{figure*}

\subsection{Overall performance}
\YeTian{Table~\ref{tbl:accuracy} reports the performance of the \Method model and representative baselines on the two datasets. The results show that \Method achieves overall mIoU that is superior or comparable to the SOTA baselines while updating only about 5\% of the features, thereby substantially reducing the computational and memory overhead of online updates. This makes \Method particularly suitable for deployment on resource-constrained devices. In practical deployment, we expect the model to generalize well to unseen environments and to be rapidly updated with as few iterations as possible. To better reflect this requirement, we set the number of online adaptation epochs to 10, which is a commonly adopted configuration for adaptive updates on edge devices.

For detailed results, on SemanticKITTI, \Method achieves 54.0\% mIoU while updating only 5\% of the features, outperforming SALUDA (49.1\%) by 4.9 percentage points and approaching the 56.9\% performance of CENET~\cite{cheng2022cenet}. On the larger and more diverse nuScenes benchmark, CENET~\cite{cheng2022cenet} attains only 54.5\% and SALUDA 65.2\%, whereas \Method reaches 61.6\% while still updating only 5\% of the data, significantly surpassing CENET~\cite{cheng2022cenet} and approaching SALUDA; meanwhile, \Method-full achieves the best mIoU of 67.3\%. Overall, \Method significantly reduces the cost of online updates while still maintaining or even surpassing the accuracy of existing SOTA methods, providing a more practical solution for continuous deployment and rapid adaptation in real-world scenarios.

Furthermore, Fig. \ref{fig:epoch_iou} shows that \Method attains a substantially higher initial mIoU than the two baselines before updates, indicating stronger adaptability in new scenes. As the number of training epochs increases, its performance curve is smooth and monotonically increasing, demonstrating a stable and reliable convergence process. In contrast, both baselines exhibit pronounced oscillations on both datasets: CENET frequently undergoes large fluctuations between adjacent epochs and, on nuScenes, even drops sharply from 0.528 at the 3rd epoch to 0.4 at the 4th epoch, making it difficult to maintain a stable optimization trajectory. Although SALUDA shows an overall upward trend, it repeatedly degrades on SemanticKITTI and oscillates between 0.456 and 0.491 in the later stage, suggesting that its gains are not robust, and its final performance also lags behind \Method. Therefore, compared with the existing representative methods CENET and SALUDA, \Method can better adapt to new scenarios and achieve rapid and stable updates.}

\subsection{Efficiency Comparison}
To compare the training efficiency of all methods, we first evaluate their Frames Per Second (FPS) on an NVIDIA RTX 4090 GPU~\cite{4090}.
As illustrated in Fig.~\ref{fig:ALL_FPS}, the results demonstrate the superior performance of \Method in training speed over all SOTA approaches. 
Compared to CENET~\cite{cheng2022cenet} and SALUDA~\cite{michele2024saluda}, \Method delivers 2.25$\times$ and 13.8$\times$ speedups, respectively, underscoring its effectiveness for latency-sensitive and resource-constrained deployments in real-world applications.
Moreover, with buffer selection, \Method reaches 84.50 FPS, nearly double the 43.48 FPS achieved by HyperLiDAR$-\textrm{full}$ on the full dataset.
\YeTian{Then, we further evaluate the training performance of \Method on resource-constrained edge devices using FSL-HDnn chip~\cite{yang2024fsl}, which is a customized HDnn ASIC designed for edge HDC workloads that can improve the efficiency of high-dimensional computations. 
Experimental results show that \Method reaches 16.3 FPS on FSL-HDnn chip. 
This highlights that by using dedicated hardware accelerators, \Method can also achieve high adaptation throughput on edge devices, emphasizing its potential for real-world deployments.}

\subsection{Impact of Buffer Size}


In streaming LiDAR scenarios, each point cloud frame is extremely dense. If all points are always used for training during online updates, the computational cost becomes substantial and is particularly prohibitive for edge devices with limited memory and computing resources. 
To address this issue, we design a buffer selection module with an adaptive update mechanism: under a fixed buffer capacity, it selectively retains newly arriving data such that samples with poorer historical performance or higher uncertainty are preferentially stored and reused. In this way, the data volume can be significantly reduced while preserving as much information as possible that is most beneficial for model updates.
Fig.~\ref{fig:buffer_selection} illustrates the efficiency–accuracy trade-off of HyperLiDAR under different buffer ratios ($k\%$). We observe that when only $5\%$ of the data is used for online updates, the FPS increases from approximately $46.15$ (with $100\%$ of the data) to $85.71$, i.e., nearly a twofold speed-up, while the mIoU remains at $0.54$, which is only about $0.017$ lower than the result obtained with $100\%$ of the data. This improvement is mainly attributed to the adaptive strategy of the proposed buffer selection module. 
As the buffer ratio increased from $5\% to $20\%, mIoU rose to $0.547$, while the FPS still maintained a relatively high processing rate of $66.67$.
These results demonstrate that the proposed buffer selection mechanism substantially reduces the computational cost of online updates with only a little loss in accuracy, effectively alleviating the burden caused by redundant point cloud data and making HyperLiDAR more suitable for efficient and stable adaptive updates in resource-constrained real-world deployment settings.

\section{Conclusion}
\label{sec:conclusion}
LiDAR segmentation is critical for understanding 3D environments and enabling spatial artificial intelligence. However, prior works have not addressed the practical bottleneck of efficient execution after deployment on resource-constrained devices. In this paper, we propose a novel LiDAR segmentation approach, \Method, designed for fast and accurate post-deployment adaptation at the edge. The method leverages the parallelism of HDC and lightweight training to effectively segment LiDAR scans. To further reduce computational overhead and improve training efficiency, \Method introduces a buffer selection strategy that accelerates training without significantly sacrificing accuracy. Experimental results show that our method outperforms SOTA LiDAR segmentation approaches while providing up to 13.8$\times$ acceleration in training.
In future work, we will further explore extending \Method to more challenging real-world settings, including label-scarce adaptation settings and lifelong learning scenarios with continuous updates.

\begin{acks}
This work has been funded in part by NSF, with award numbers \#2112665, \#2112167, \#2003279, \#2120019, \#2211386, \#2052809, \#1911095 and in part by PRISM and CoCoSys, centers in JUMP 2.0, an SRC program sponsored by DARPA.
\end{acks}

\newpage

\bibliographystyle{ACM-Reference-Format}
\bibliography{refs}

\end{document}